\documentclass[fleqn,10pt]{wlscirep}
\usepackage[utf8]{inputenc}
\usepackage[T1]{fontenc}
\title{Towards Privacy-aware Mental Health AI Models: Advances, Challenges, and Opportunities}

\author[1,2]{Aishik Mandal}
\author[3,4,*]{Tanmoy Chakraborty}
\author[1,2,*]{Iryna Gurevych}
\affil[1]{Ubiquitous Knowledge Processing Lab (UKP Lab)\\
Department of Computer Science and Hessian Center for AI (hessian.AI)\\
Technische Universität Darmstadt}
\affil[2]{National Research Center for Applied Cybersecurity ATHENE, Germany}
\affil[3]{Department of Electrical Engineering, Indian Institute of Technology Delhi, India}
\affil[4]{Yardi School of Artificial Intelligence, Indian Institute of Technology Delhi, India}

 \affil[*]{Corresponding author: Tanmoy Chakraborty (tanchak@iitd.ac.in) and Iryna Gurevych (iryna.gurevych@tu-darmstadt.de)}

\begin{abstract}

Mental health disorders create profound personal and societal burdens, yet conventional diagnostics are resource-intensive and limit accessibility. Advances in artificial intelligence, particularly natural language processing and multimodal methods, offer promise for detecting and addressing mental disorders, but raise critical privacy risks. This paper examines these challenges and proposes solutions, including anonymization, synthetic data, and privacy-preserving training, while outlining frameworks for privacy–utility trade-offs, aiming to advance reliable, privacy-aware AI tools that support clinical decision-making and improve mental health outcomes.

\end{abstract}
\begin{document}

\flushbottom
\maketitle
%
%
\thispagestyle{empty}


\section*{Introduction}

Mental disorders are a leading cause of disability worldwide, with profound societal, economic, and personal consequences. Early and accurate diagnosis is critical, yet current clinical practices, relying on self-reported questionnaires and time-intensive clinical interviews, face scalability challenges due to the global shortage of trained therapists. Consequently, many individuals remain undiagnosed or untreated. Routine completion of post-session questionnaires also imposes an additional burden on patients.


These limitations have spurred interest in AI-based systems designed to automate diagnosis and assist in therapy. In clinical settings, therapists rely on a range of multimodal cues: verbal, visual, and acoustic, to assess mental health conditions. For instance, depression is associated with characteristic facial expressions \cite{slonim2023facing,5349358,SCHERER2014648}, speech prosody \cite{cummins2015review,5349358,SCHERER2014648}, and semantic patterns \cite{chim-etal-2024-overview}. Anxiety often manifests as difficulty maintaining eye contact \cite{langer2017social,app122312298}. Speech features are instrumental in detecting Post-Traumatic Stress Disorder (PTSD) \cite{kathaneffect,HU2024859}, while both speech and facial features are valuable for identifying Bipolar Disorder \cite{bipolar_speech,gilanie2024robust}. Accordingly, multimodal AI models \cite{mult_ai_survey} that integrate text, audio, and video data have shown promise in improving diagnosis and support therapists in managing mental health conditions. Recent approaches have combined acoustic features from spectrograms with visual data from convolutional neural networks \cite{mult-ai-dep}, or incorporated facial action units, gaze, and head pose extracted using OpenFace \cite{openface} along with textual data \cite{sadeghi2024harnessing}. Others utilize audio descriptors (e.g., MFCCs, eGeMAPS) together with facial landmarks, gaze, head pose and AUs to improve classification of depression and bipolar disorder \cite{mult-ai-bd-dep}. Fusion methods such as adapter-based fusion \cite{mult-ai-sz}, cross-modal attention based fusion \cite{cross-mod-attention} and  early fusion mechanisms \cite{mult-ai-anx-dep} have further improved predictions of mental disorders.

Despite these advances, the development of multimodal AI for mental health remains hindered by privacy concerns. Training such models requires sensitive therapy recordings containing identifiable speech and facial data, subject to regulations including the General Data Protection Regulation (GDPR) \cite{gdpr} and the Health Insurance Portability and Accountability Act (HIPAA) \cite{act1996health}. These data cannot be publicly released, and many patients are reluctant to participate without strong privacy guarantees. As a result, existing datasets are often small and biased, limiting model performance and generalizability. Moreover, releasing model weights trained on sensitive data may inadvertently leak private information through memorization or inference attacks, posing serious ethical and legal risks \cite{membership-inference-attack, embedding-data-leakage,nn-memorise, fine-tune-data-leak}. Such privacy breaches could expose patients’ identities or enable impersonation, potentially worsening their mental health. These challenges significantly hinder the development and deployment of mental health AI models in real-world applications.


To address these challenges, privacy-preserving techniques are increasingly being explored. These techniques fall into two broad categories: data privacy, which involves anonymizing while retaining clinical relevance\cite{lstm_pii,lstm_bert_pii,gpt-4-deid,speech-pii-flechl,VP2022,tomashenko2024voiceprivacy,human_attribute_privacy} or synthetic data generation \cite{chen-etal-2023-soulchat,PTSD_synth,patient-psi,lee-etal-2024-cactus,chu2024syntheticpatientssimulatingdifficult}; and model privacy, which focuses on training algorithms robust to privacy breaches, using methods such as differential privacy (DP) \cite{dp-sgd,context-aware-dp-lm,plant-etal-2021-cape,yu2022differentially,kerrigan-etal-2020-differentially,aud-dp,bu2022differentially}. However, these protections often reduce model utility, underscoring the need for rigorous evaluation metrics for both privacy \cite{re-id,pii_poor_gen,VP2022,tomashenko2024voiceprivacy,unlinkability-2,human_attribute_privacy,demogarphic_face_privacy,synthetic-privacy-all,MURTAZA2023100546,privacy-gain,context-aware-dp-lm} and utility \cite{re-id,VP2022,tomashenko2024voiceprivacy,multi-speaker-anon,human_attribute_privacy,qiu-2024,zhang2024cpsycounreportbasedmultiturndialogue,Synth_Soc_1,Mehta_2024_CVPR,Mughal_2024_CVPR,context-aware-dp-lm}. Both automatic \cite{qiu-2024,zhang2024cpsycounreportbasedmultiturndialogue} and human \cite{chen-etal-2023-soulchat,Mughal_2024_CVPR,Mehta_2024_CVPR,VP2022} evaluations are used to quantify the privacy-utility trade-off and to guide the development of effective, safe systems.

\begin{figure}[!ht]
\centering
\includegraphics[width=\linewidth]{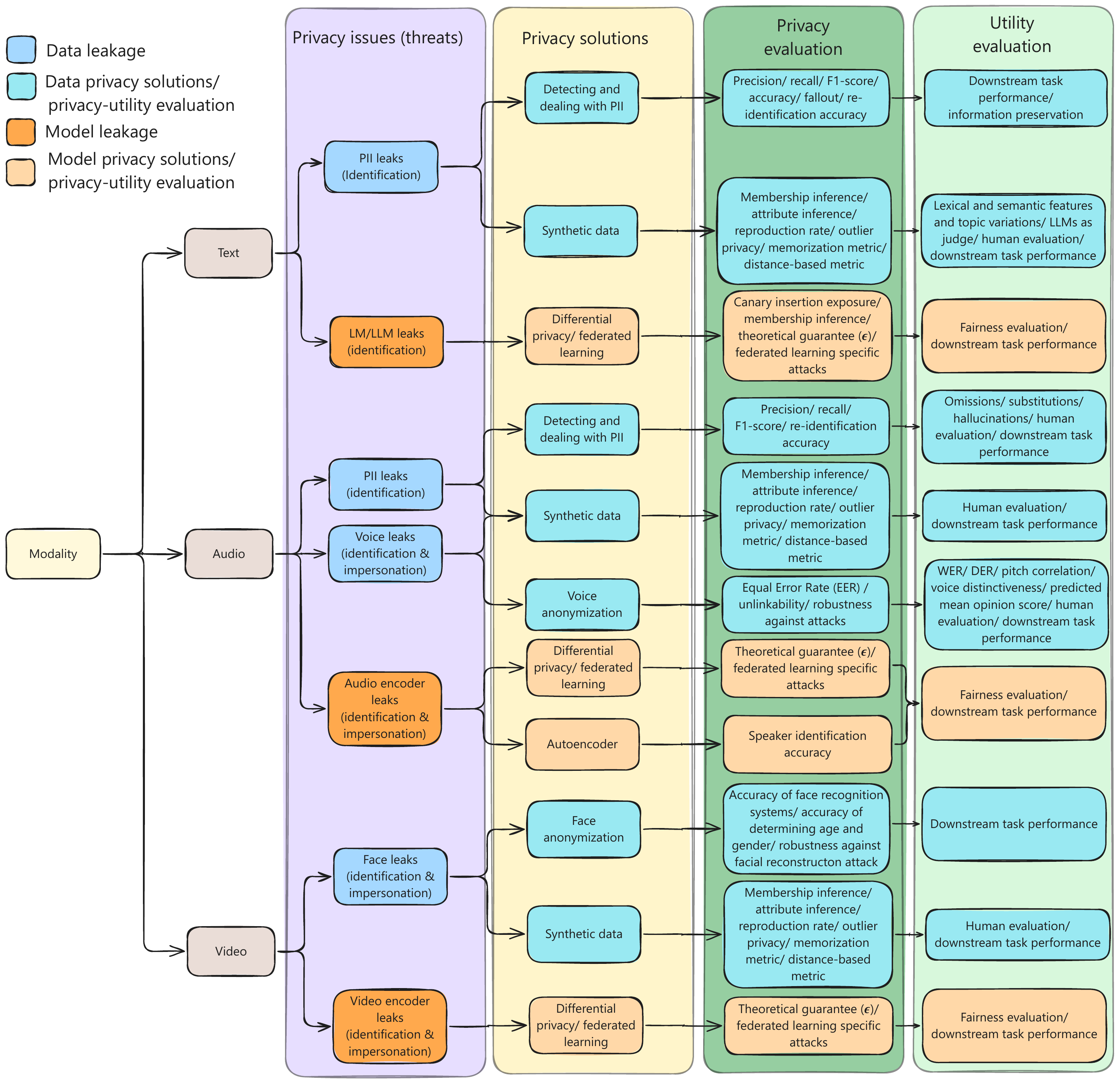}
\caption{\textbf{Overview of privacy challenges, potential solutions and privacy-utility evaluations in Mental Health AI.} Potential solutions to address current privacy challenges and threats across modalities in mental health dataset creation, as well as in the development and evaluation of mental health AI models, to determine the privacy-utility trade-offs of the solutions.}
\label{fig:schematic}
\end{figure}

In summary, multimodal AI models hold significant potential to assist therapists and make mental illness diagnoses more accessible. However, privacy issues limit the availability of suitable datasets and, consequently, the development of robust models for real-world deployment. We discuss these privacy issues and explore potential solutions that can ensure privacy in mental health datasets and models. Additionally, we explore evaluation methods to analyze the privacy-utility trade-offs of these solutions. Figure \ref{fig:schematic} presents a schematic diagram summarizing the discussion. Finally, we recommend a privacy-aware pipeline for data collection and model training and outline future research directions to support the development of such a pipeline.

\section*{Current Privacy Issues}

Current privacy issues in mental health datasets and models include the risk of private information leaking from both data and models to malicious actors. Privacy leakage from data prevents the public release of datasets, while leakage from models restricts the sharing of trained model weights.

\subsection*{Private information leakage from datasets}

Privacy leakage from data includes personally identifiable information (PII) present in text transcripts and audio recordings of therapy sessions. Additionally, the audio-visual therapy recordings reveal the voices of patients and therapists, as well as the faces of patients. Malicious actors can also exploit the extracted audio and video features used in mental health diagnosis models to infer sensitive attributes and PII, such as the patient’s age and gender or to reconstruct their voices and faces.

\paragraph{PII leakage.} In the EU, the GDPR protects PII like name, identification number, location data, online identifier, or factors specific to their physical, physiological, genetic, mental, economic, cultural, or social identity. Similarly, in the US, HIPAA protects individually identifiable health information, including details such as an individual’s name, address, birth date, Social Security Number, and records of their past, present or future physical or mental health conditions. Many of these types of personal information are frequently discussed in therapy sessions. While such information can be identified and removed in structured data formats like tables, therapy sessions often involve detailed personal narratives, which can inadvertently reveal sensitive information. As a result, textual transcripts and speech recordings of therapy sessions often contain PII that could be used to identify a patient. Even with anonymization of PII, they can still show identification vulnerabilities through the use of other public datasets \cite{cross-dataset-identity}.


\paragraph{Voice from audio.} Speech data are classified as personal data under GDPR because they can reveal sensitive information about the speaker, including their identity, age, gender, health status, personality, racial or ethnic origin and geographical background \cite{NAUTSCH2019441}. Features commonly used in mental health models such as MFCCs, pitch, and Mel-spectrograms can unintentionally encode sensitive information such as a person's age and gender \cite{speech_gender_age}. Studies have shown that MFCCs can support speech reconstruction \cite{mfcc_speech_reconstruct}, while learned embeddings like Wav2Vec can be used for voice conversion \cite{wav2vec-vc}, enabling impersonation of both patients and therapists.


\paragraph{Face from video.} Mental health models typically utilize facial features extracted from deep encoder models such as ResNet \cite{He_2016_CVPR} or facial landmarks, obtained through tools like OpenFace \cite{openface} for behavior, expression, and gaze analysis. However, features derived from deep learning models \cite{Schroff_2015_CVPR,alexnet} are susceptible to reconstruction attacks, enabling adversaries to recover facial identities \cite{Dosovitskiy_2016_CVPR,template_reconstruct}. Even lower-level features like facial landmarks can be used to reconstruct approximate facial geometry \cite{landmark_reconstruct}.


\subsection*{Private information leakage from models}

Trained models are often susceptible to leaking training data when subjected to attacks, such as membership inference attacks \cite{membership-inference-attack}, from malicious actors. This is particularly concerning for models trained on mental health datasets, in which case leaks will lead to revealing private data. Mental health datasets are generally small and have a higher prevalence of rare data points. Embedding models are particularly vulnerable to leaking such infrequent training data inputs \cite{embedding-data-leakage}. Neural networks also memorize such unique training data points, which can then be extracted from the trained models \cite{nn-memorise}. Moreover, in text, private data can be leaked through context \cite{context-aware-dp-lm}. This is especially true for the mental health domain, where discussing life events can indirectly leak private data. Models are also prone to exposing user information contained in the data used for fine-tuning \cite{fine-tune-data-leak}. This poses a significant privacy challenge to releasing models trained or fine-tuned on mental health datasets, as they may inadvertently memorize and disclose sensitive patient information. 



\section*{Threats}

The leakage of a mental health patient’s private information, such as their voice or face, can lead to identification, social stigma, and exploitation. This includes risks of defamation, blackmail through deepfakes, impersonation, and misuse of biometrics, which could worsen the patient’s mental health condition.

\paragraph{Identification.}


Private data leakage from mental health datasets can lead to patient identification and public exposure of their mental health records, resulting in workplace discrimination, social isolation, and blackmail, further aggravating their mental condition. Sensitive information, such as age, address, and gender, revealed during therapy or extracted from audio and video recordings, can uniquely identify most Americans \cite{pii_identification}. Large Language Models (LLMs) trained on therapy data are prone to privacy breaches, leaking such information \cite{membership-inference-attack, embedding-data-leakage, nn-memorise, fine-tune-data-leak}. Voice data can be exploited for identification via speaker verification systems \cite{identification_asv, identification_asv_3}, while video data may reveal faces, enabling identification through face recognition \cite{identification_face_reco, Huber_2024_WACV}.


\paragraph{Impersonation.}


The leakage of voice and video data from mental health datasets enables malicious agents to impersonate patients through deepfakes, which can be audio, video, or audio-visual. Audio deepfakes use a person’s voice for false speech or impersonation via voice conversion, text-to-speech, and replay attacks \cite{speech_deepfake_1,speech_deepfake_4,KIETZMANN2020135,vid_deepfake}. Impersonation attacks by humans mimicking speech traits also pose a risk \cite{impersonation_voice}.  
Video deepfakes manipulate faces and bodies using reenactment, video synthesis, and face swaps \cite{TOLOSANA2020131,KIETZMANN2020135,vid_deepfake,speech_deepfake_4}, while audio-visual deepfakes combine voice and appearance \cite{av_deepfake,KIETZMANN2020135}. Deepfakes can be exploited for fraud, blackmail, harassment, identity theft, and other malicious activities \cite{MUSTAK2023113368}, causing severe psychological distress and worsening patients' mental health.




\section*{Addressing the Privacy Issues}

Privacy concerns in mental health datasets can be addressed through data anonymization or by generating synthetic data derived from real datasets. Data anonymization involves removing PII from therapy transcripts and audio recordings, as well as applying voice and face anonymization techniques while preserving features crucial for mental health diagnosis. An alternative approach is the creation of synthetic data that mimics the real dataset without exposing specific patient attributes. Homomorphic encryption can also be used for data protection \cite{tee}; however, it demands significant computational resources, making it impractical in many cases \cite{tee-bad}. Privacy issues arising from models trained on mental health datasets leaking patient information can be mitigated using privacy-aware training methods.

\subsection*{Data anonymization}


Data anonymization involves removing PII in transcripts and audio recordings, voice anonymization in audio recordings and face anonymization in video recordings of therapy sessions to prevent identification and impersonation threats. Below, we outline approaches for anonymizing textual, audio, and visual data to ensure privacy while retaining essential information for mental health diagnosis and discuss their limitations.

\paragraph{Text anonymization by detecting and removing PII.}

The presence of PII in therapy transcripts, such as names, addresses, and dates, poses significant privacy risks. Common approaches employ Named Entity Recognition (NER) models to detect and replace PII with synthetic tokens \cite{ner_de_id,lstm_pii}, yet these systems struggle with generalization across diverse contexts \cite{pii_poor_gen,lstm_bert_pii,kim-etal-2024-generalizing} and require domain-specific fine-tuning \cite{gpt-4-deid}. Existing efforts, such as dataset augmentation with synthetic data \cite{kim-etal-2024-generalizing,lstm_bert_pii} and methods tailored to spoken language \cite{speech_pii_augment}, offer incremental improvements. Yet, these advances remain largely confined to Electronic Health Records (EHRs) and do not tackle the unique challenges of therapy sessions where sensitive disclosures unfold dynamically over multiple interactions. Even approaches targeting spoken data\cite{speech_pii_augment} rely on overly simplistic setups like passage reading, failing to capture the nuances of actual patient-therapist exchanges. Moreover, therapy conversations often contain implicit and contextualized disclosures of private information that simple NER-based systems fail to capture \cite{baroud-etal-2025-beyond}. Recent work has explored large language models (LLMs) for de-identification \cite{gpt-4-deid}, but their real-world utility is deeply flawed as the reliance on API-based models involves sending sensitive data to proprietary models which can be used to train these models \cite{balloccu-etal-2024-leak} thus compromising privacy.

A particularly glaring gap is the absence of privacy frameworks designed for longitudinal therapy data. Unlike standalone medical records, therapy transcripts span multiple sessions, making cross-session privacy breaches a critical yet overlooked concern. The failure of current methods to provide theoretical DP guarantees only increases the problem, raising fundamental questions about the reliability of anonymization techniques used in mental health AI models. If we are to responsibly advance AI-driven mental health support, the field must move beyond ad hoc anonymization techniques and develop robust, theoretically grounded privacy mechanisms tailored for real-world conversational therapy data.

\paragraph{Audio anonymization by addressing PII in speech data.}

Audio anonymization typically relies on Automatic Speech Recognition (ASR) followed by NER-based PII redaction. Simple approaches such as replacing PII with silence \cite{cohn-etal-2019-audio}, beeps, or white noise \cite{Veerappan2024} compromise naturalness and degrade usability. More advanced techniques replace PII with synthetic content from the same category, and synthesize it into speech \cite{Veerappan2024}. However, this approach fundamentally alters the entire audio rather than making targeted modifications.
A more refined strategy would involve selectively replacing PII segments while preserving the rest of the speech, such as splicing in matching audio fragments\cite{speech-pii-flechl}. However,  these methods are yet to be validated on longitudinal therapy data, leaving them vulnerable to cross-session linkages. Like their textual counterparts, current audio anonymization methods lack formal privacy guarantees, casting doubt on their effectiveness for real-world deployment.


\paragraph{Voice anonymization for speaker privacy.}

Voice anonymization methods, while effective for protecting speaker identity in speech and emotion recognition tasks\cite{VP2022,tomashenko2024voiceprivacy}, remain inadequate for mental health applications. Existing techniques modify x-vectors\cite{OHNN}, pitch, and bottleneck features\cite{champion:hal-03753746,shamsabadi:hal-03588932} to obscure speaker identity while preserving intelligibility and emotional cues. Some even offer theoretical DP guarantees \cite{shamsabadi:hal-03588932}. However, these methods have not been evaluated for their ability to retain speech markers essential for mental health diagnosis, raising concerns about their applicability. %
An even greater limitation is their focus on Single Speaker Anonymization (SSA), whereas therapy sessions involve dynamic conversations between therapists and patients, necessitating Multi-Speaker Anonymization (MSA). While Miao et al. \cite{multi-speaker-anon} introduced an MSA benchmark, real-world therapy scenarios involve overlapping speech and stronger attack models, which remain unaddressed. Without rigorous testing on mental health datasets and the development of DP-compliant MSA techniques, current methods fall short of ensuring privacy while preserving diagnostic speech features.

\paragraph{Face anonymization for visual privacy.}

Face anonymization in video recordings remains an unsolved challenge, especially for mental health applications. While tools like Face-Off\cite{face-off}, LowKey\cite{lowkey}, Foggysight\cite{foggysight}, and FAWKES\cite{fawkes} attempt to obfuscate faces in images, they often fail against adaptive face recognition systems\cite{perturb_faces}. AI stylization\cite{vid_anon} offers a promising alternative by preserving emotional expressions, but its effectiveness in real-world settings remains uncertain.  
Frame-by-frame image anonymization is computationally expensive in video contexts. More efficient video-specific tools like FIVA \cite{Rosberg_2023_ICCV} offer promise but introduce demographic biases \cite{demogarphic_face_privacy} and may leak sensitive attributes such as age or gender \cite{human_attribute_privacy,gender_leak_face_obfus}. Even methods with DP guarantees \cite{vid-dp,img-dp} remain unproven in clinical applications. Furthermore, most visual anonymization overlooks non-facial behavioral cues such as body language which may reveal gender \cite{gender_leak_face_obfus}. Thus, video anonymization must go beyond facial modifications to ensure privacy without erasing clinically relevant behavioral cues.


\subsection*{Synthetic data generation}
Synthetic data, generated using AI models, mirrors real data but does not belong to actual individuals, ensuring privacy. It offers a solution to data scarcity and diversity challenges in mental health datasets, enabling effective AI training while protecting sensitive information.

\paragraph{Synthetic text generation.} Textual synthetic data generation for therapy transcripts remains limited by the shortcomings of current methods. Approaches such as prompting LLMs\cite{PTSD_synth, lozoya-etal-2024-generating}, converting single-turn psychological Q\&A or counseling reports into multi-turn dialogues\cite{chen-etal-2023-soulchat,qiu2024smilesingleturnmultiturninclusive,zhang2024cpsycounreportbasedmultiturndialogue}, and role-playing setups with integrated cognitive knowledge \cite{patient-psi,lee-etal-2024-cactus} offer promising directions. Some even provide DP guarantees \cite{yue-etal-2023-synthetic,nahid2024safesynthdp}.  
However, these approaches typically produce short dialogues (averaging 17 turns) and fail to replicate the complexity and duration of real therapy sessions. LLMs struggle with coherence, memory, and instruction retention over extended conversations \cite{multi-turn-struggle}, limiting their clinical utility. Without a robust multi-turn generation framework that can sustain long, contextually consistent dialogues, synthetic therapy data will remain artificial and clinically inadequate.


\paragraph{Synthetic multimodal data generation.} Given the superior performance of multimodal models in mental health diagnosis, synthetic multimodal data generation is critical. Current methods employ sequential pipelines to generate speech, gestures, and video from text \cite{Mehta_2024_CVPR,li2024styletalker,Mughal_2024_CVPR,Ng_2024_CVPR, chu2024syntheticpatientssimulatingdifficult}, but suffer from noise accumulation, and computational inefficiencies that hinder real-world applicability.
%
%
A more fundamental issue is the lack of psychiatric knowledge in these methods. Moreover, these methods remain constrained to short clips, while real therapy sessions span 30 to 60 minutes and often extend across multiple sessions. To advance synthetic data generation in mental health, future methods must support long-form, contextually coherent, multimodal generation grounded in clinical knowledge. Without such innovations, synthetic data will remain insufficient for developing clinically robust AI models.


\subsection*{Privacy-aware training}

Privacy-aware training methods are essential for developing AI models in mental health, ensuring that private and sensitive data is protected in trained models while maintaining model utility.

\paragraph{Differential Privacy.} DP\cite{DP} offers strong theoretical privacy guarantees and is widely used for training privacy-aware models through Differentially-Private Stochastic Gradient Descent (DP-SGD) \cite{dp-sgd}. However, its application in language modeling remains problematic due to significant performance degradation \cite{kerrigan-etal-2020-differentially}. DP techniques have been extended to conformer-based audio encoders\cite{aud-dp} and ResNet models for image and video data\cite{bu2022differentially}, as well as differentially private fine-tuning methods for sensitive datasets like those in mental health\cite{kerrigan-etal-2020-differentially,yu2022differentially}.

Yet, these efforts remain incomplete. DP methods have not been adapted for multimodal models that integrate text, audio, and video, leaving them vulnerable to cross-modal information leakage. This is concerning for multimodal mental health applications. Furthermore, therapy sessions unfold over multiple interactions, making context-aware DP essential. While recent work attempts to incorporate contextual information \cite{context-aware-dp-lm,plant-etal-2021-cape}, their efficacy remains untested on longitudinal mental health data. Without specialized DP techniques that account for long-range context dependencies in therapy data, existing privacy mechanisms fall short of ensuring meaningful protection in real-world mental health applications.


\paragraph{Federated learning.} Federated Learning (FL)\cite{fl} is widely promoted as a privacy-preserving training method, enabling decentralized model training without directly sharing patient data. Although conceptually attractive for mental health settings, FL alone offers weak privacy guarantees and remains susceptible to gradient leakage attacks \cite{fl-attack1,fl-attack2,fl-attack3}, as sensitive information can still leak through model gradients and local weights\cite{NAGY2023110475}. 
%
%
Combining FL with Local Differential Privacy (LDP) improves security \cite{NAGY2023110475,dp-fl1}, but often results in significant performance loss on realistic medical datasets, small datasets or large models \cite{dp-fl-benchmark}, limiting practical adoption in mental health applications. Without substantial improvements in both privacy protection and model utility, deploying FL-based approaches in real-world mental health datasets remains premature.


\paragraph{Confidential computing with Trusted Execution Environment (TEE).} Confidential computing aims to safeguard the data during processing. It loads the data and the model in a TEE where they are protected from unauthorized access and modification \cite{tee}. However, it requires more computational resources and special hardware, limiting its wide-spread usage \cite{tee-bad}.

\paragraph{Autoencoders for privacy preservation.}
Autoencoders are commonly used in speech models to extract latent representations that retain mental health relevant features while obfuscating speaker identity \cite{ravuri,pranjal}. While promising in speech-based tasks, their use in multimodal therapy data remains largely unexplored.



\section*{Evaluating Privacy-aware Alternatives}

Ensuring privacy in AI models for mental health diagnosis is essential to protect patient confidentiality. However, this often comes at the cost of reduced performance in downstream diagnostic tasks. This section discusses methodologies that should be used for evaluating privacy-utility trade-offs across three key areas: data anonymization, synthetic data generation, and privacy-aware training. Without standardized, context-specific benchmarks, current approaches risk being either too lax to ensure real privacy or too restrictive to be clinically useful.

\subsection*{Data anonymization}

Data anonymization techniques focus on removing or masking private information across text, audio, and video modalities. Effective anonymization should minimize privacy risks while preserving the diagnostic utility of the data. Evaluation can be categorized into privacy and utility metrics.

\paragraph{Privacy evaluation.} Effective evaluation of privacy in mental health AI models must go beyond standard metrics such as precision, recall, or F1-score for NER-based PII removal techniques, which focus narrowly on explicit identifiers \cite{SANCHEZ2014189,lstm_pii,pii_poor_gen,kim-etal-2024-generalizing,gpt-4-deid,lstm_bert_pii}. Similarly, audio anonymization methods are often evaluated using PII detection metrics \cite{cohn-etal-2019-audio,Veerappan2024,speech-pii-flechl}, but these overlook deeper privacy threats present in mental health data like indirect leaks. 
Advanced privacy assessment requires adversarial re-identification models, including LLM-based techniques capable of detecting indirect leaks across longitudinal therapy sessions \cite{re-id}. Privacy risks must also be evaluated in relation to external datasets, where residual identifiers can re-emerge \cite{cross-dataset-identity}. Voice anonymization techniques must be tested using Equal Error Rate (EER) \cite{VP2022,tomashenko2024voiceprivacy,OHNN,champion:hal-03753746,shamsabadi:hal-03588932}, False Accept Rate (FAR) \cite{multi-speaker-anon} in Automatic Speaker Verification (ASV) systems, unlinkability \cite{shamsabadi:hal-03588932,unlinkability-2} and robustness against attack models \cite{VP2022,tomashenko2024voiceprivacy,OHNN}. For video, facial anonymization must be assessed for resistance to face recognition \cite{face-off,lowkey,foggysight,fawkes,Rosberg_2023_ICCV} and reconstruction\cite{Rosberg_2023_ICCV}, attribute leakage (e.g., age, gender) \cite{human_attribute_privacy}, and demographic bias \cite{demogarphic_face_privacy}. Body language obfuscation should also be evaluated for gender inference risks \cite{gender_leak_face_obfus}.

Crucially, privacy assessments remain modality-specific, ignoring the potential for cross-modal leakage, a critical issue in multimodal mental health datasets. True privacy evaluation must involve multimodal re-identification models that account for interaction between text, audio, and video.


\paragraph{Utility evaluation.} Privacy in mental health AI models is meaningless if it comes at the expense of diagnostic accuracy. Any anonymization method must preserve clinically relevant information while ensuring privacy, making utility evaluation just as critical as privacy assessment.  

For text anonymization, utility should be evaluated on the preservation of clinically relevant information \cite{SANCHEZ2014189,re-id} and model performance on downstream diagnostic tasks. PII removed audio utility assessments should include substitution, hallucination, and omission rates \cite{Veerappan2024}, particularly in speech segments crucial for mental health analysis, alongside human evaluations of naturalness and style. Models trained on anonymized audio should finally be tested for downstream task performance. 
Voice anonymization should be evaluated via intelligibility (Word Error Rate) \cite{VP2022,tomashenko2024voiceprivacy,OHNN,multi-speaker-anon}, emotion preservation (emotion recognition accuracy) \cite{tomashenko2024voiceprivacy}, pitch correlation \cite{VP2022}, voice diversity (Gain of Voice Distinctiveness) \cite{VP2022,OHNN}. Human evaluations (naturalness, intelligibility) \cite{VP2022} and automatic metrics like Predicted Mean Opinion Score (PMOS) \cite{multi-speaker-anon} provide further insight. In case of multi-speaker anonymization situations, such as therapy sessions, the Diarization Error Rate (DER) \cite{multi-speaker-anon} should also be evaluated to ensure that speaker separation remains intact. Finally, models using anonymized voice data should be tested on downstream mental health applications.
For face anonymization, utility should be tested by evaluating the performance of anonymized videos in downstream tasks such as emotion detection and mental health diagnosis \cite{human_attribute_privacy,vid_anon}.

However, evaluating each modality separately overlooks the bigger issue: mental health diagnosis relies on the interplay between text, audio, and video. A true test of an anonymization method’s viability is the performance of a model trained on anonymized versions of all three modalities on downstream mental health tasks. Without this, utility assessments remain incomplete, and privacy-preserving methods risk being clinically irrelevant.


\subsection*{Synthetic data generation}

Synthetic data is generated by models trained on real-world data and may inadvertently reveal sensitive information if the models overfit, particularly when the real-world dataset is small (which is the case for most mental health datasets). Overfitting increases the risk of privacy violations, while overly generic synthetic data can reduce utility. Thus, evaluating privacy-utility trade-offs in synthetic data generation is crucial.

\paragraph{Privacy evaluation.} Synthetic data must be rigorously tested for privacy vulnerabilities. Standard assessments should measure robustness against membership inference and attribute inference attacks \cite{goncalves2020generation,privacy-gain,MURTAZA2023100546,synthetic-privacy-all}, as these attacks can reveal whether a specific individual’s data was part of the training set. The challenge is even greater in mental health datasets, where data is not only sparse but also highly diverse, making models more prone to overfitting on outliers. Thus, privacy gain \cite{privacy-gain} and outlier similarity \cite{MURTAZA2023100546} should be key evaluation criteria.

Beyond individual data points, synthetic datasets must also be scrutinized for direct content reproduction. Measuring reproduction rate \cite{MURTAZA2023100546} helps detect whether any real therapy sessions have been copied verbatim into the generated dataset. Additional privacy metrics like memorization coefficients \cite{mem-coeff}, $\epsilon$-identifiability \cite{eps-identifiability}, and distance-based metrics \cite{MURTAZA2023100546,synthetic-privacy-all} should be employed to ensure a more comprehensive evaluation. Without these safeguards, synthetic data may introduce more risks than solutions.

\paragraph{Data quality and utility evaluation.} Synthetic data quality should be assessed through faithfulness (similarity to real-world data) and diversity (lexical, semantic, and topic variation) \cite{qiu2024smilesingleturnmultiturninclusive,qiu-2024}. This should be complemented by mental health professionals performing expert evaluations on aspects like naturalness, empathy, helpfulness, professionalism, comprehensiveness, authenticity, and safety \cite{chen-etal-2023-soulchat,zhang2024cpsycounreportbasedmultiturndialogue}. For multimodal synthetic data, metrics like coherence, contextual plausibility,  overall naturalness of generated speech and gesture should also be measured \cite{Mehta_2024_CVPR,Mughal_2024_CVPR}. For large-scale evaluations, LLMs provide a scalable alternative to rate generated data on the same aspects as expert evaluations \cite{chen-etal-2023-soulchat,zhang2024cpsycounreportbasedmultiturndialogue}. For multimodal evaluations, multimodal LLMs \cite{next-gpt} can be employed to assess the integration of speech, text, and video.


Ultimately, the best test of synthetic data utility lies in its real-world applicability. A crucial final evaluation step is training a multimodal model on the generated data and measuring its performance on a downstream mental health diagnosis task. Without strong predictive utility, privacy alone is insufficient.


\subsection*{Privacy-aware training}

Mental health AI models must employ privacy-aware training methods. However, such methods introduce noise, necessitating a privacy-utility evaluation to identify optimal approaches that balance privacy and utility.

\paragraph{Privacy evaluation.} To ensure privacy in mental health AI, models must be rigorously tested for information leakage. Language models should be evaluated using canary insertion and membership inference attacks \cite{context-aware-dp-lm}, while embeddings must be assessed for leakage of sensitive attributes (like location, age, gender, or identity) \cite{plant-etal-2021-cape}. For models trained with DP-SGD, the privacy guarantee is quantified by the $\epsilon$-value \cite{dp-sgd} (which determines the distance within which errors are considered to be zero in Stochastic Gradient Descent). In FL, evaluations must account for gradient and weight leakage and FL specific attacks \cite{NAGY2023110475}. Without such targeted evaluations, privacy-preserving claims remain unverified and models may remain vulnerable to inference threats.


\paragraph{Utility evaluation.}
Utility evaluation examines model performance on downstream mental health diagnosis tasks \cite{context-aware-dp-lm,plant-etal-2021-cape,kerrigan-etal-2020-differentially,yu2022differentially,bu2022differentially,ravuri,pranjal}. However, differential privacy training often exacerbates model unfairness \cite{dp-sgd-bad-performance}. Thus, utility evaluations must also consider the performance of privacy-aware models on culturally and demographically diverse mental health datasets.

\begin{figure}[!ht]
\centering
\includegraphics[width=0.95\linewidth]{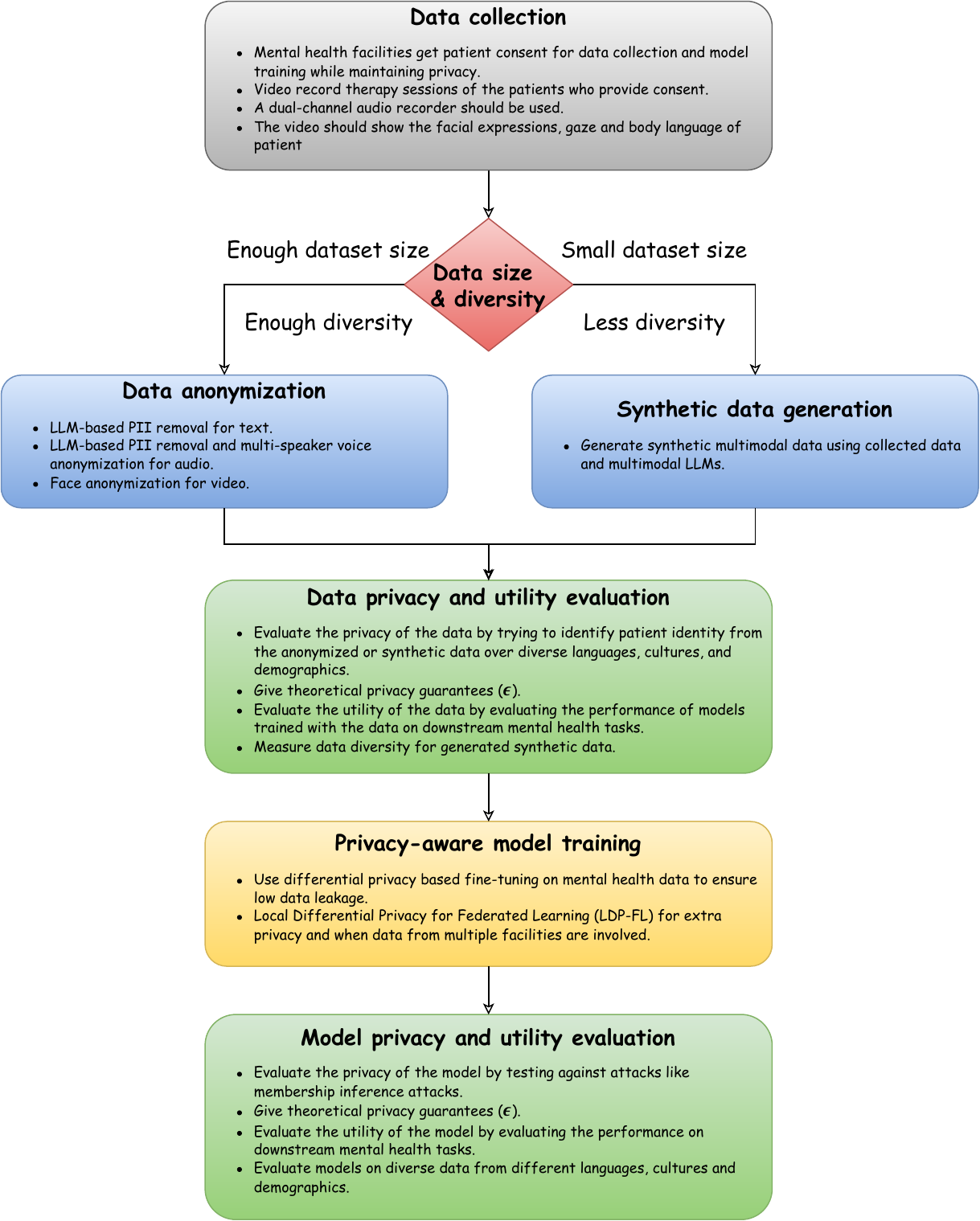}
\caption{\textbf{Proposed pipeline for developing privacy-aware mental health AI.} Our proposed pipeline for data collection and model training enables the development of privacy-aware mental health AI models, integrating safeguards to balance privacy with clinical utility.}
\label{fig:workflow}
\end{figure}

\section*{Recommendations}
Based on the advances and pitfalls of existing studies, we recommend a comprehensive workflow for developing privacy-aware mental health AI models and datasets. The workflow involves data collection, data anonymization as well as synthetic data generation, privacy-utility evaluation of the data, privacy-aware model training, and evaluation of the privacy-utility trade-off in the training process. Figure \ref{fig:workflow} shows the recommended pipeline.

\paragraph{Data collection.}
Developing privacy-preserving mental health AI systems begins with careful and ethically sound data collection. Therapy sessions must be recorded only with explicit, informed patient consent, detailing anonymization, secure storage, and exclusive research use. Approval from institutional ethics committees is essential. Audio should be captured using dual-channel recorders to separate therapist and patient voices, while video should focus on patients’ full facial and postural expressions for behavioral analysis. Transcription should be conducted locally using researchers or offline ASR tools to avoid third-party data exposure.

\paragraph{Data anonymization and synthetic data generation.}
Once collected, data must either be anonymized or replaced with synthetic data, depending on dataset scale and diversity. Small or demographically narrow datasets may require synthetic data augmentation to ensure sufficient coverage for training generalizable models. Larger, diverse datasets may benefit from direct anonymization. For anonymizing text transcripts, LLMs can be employed to detect and redact PII \cite{gpt-4-deid}. However, proprietary models cannot be used due to privacy issues. Thus, open source LLMs should be fine-tuned on augmented long-context conversational datasets for improved generalization \cite{kim-etal-2024-generalizing,speech_pii_augment,lstm_bert_pii}. Audio anonymization can be achieved by selectively replacing PII-bearing audio segments through either synthesizing matching audio segments through splicing in matching audio fragments \cite{speech-pii-flechl} or emotion-aware voice synthesis \cite{emotion-aware-voice} and voice conversion \cite{wav2vec-vc}, preserving speech naturalness. Multi-speaker anonymization methods \cite{multi-speaker-anon} should be used to maintain conversational dynamics while obscuring identity. For video, face anonymization tools such as FIVA \cite{Rosberg_2023_ICCV} should be employed, though body language must also be obfuscated to prevent gender or identity leakage. Alternatively, synthetic datasets may be generated using multimodal LLMs \cite{next-gpt} via role-playing as therapists and patients \cite{qiu-2024}, zero-shot, or few-shot prompting \cite{PTSD_synth}, ensuring no direct mapping to real individuals.

\paragraph{Privacy-utility evaluation of anonymized and synthetic data.}
Both anonymized and synthetic data must be evaluated for privacy-utility trade-offs. Privacy evaluations should include re-identification tests using adversarial models \cite{re-id}, other related datasets \cite{cross-dataset-identity} and measuring the effectiveness of techniques like multi-speaker anonymization through metrics such as EER \cite{VP2022,tomashenko2024voiceprivacy,OHNN,champion:hal-03753746,shamsabadi:hal-03588932} and FAR \cite{multi-speaker-anon} in speaker verification. Voice anonymization should also be evaluated for unlinkability \cite{shamsabadi:hal-03588932,unlinkability-2} and robustness against attacks \cite{VP2022,tomashenko2024voiceprivacy,OHNN}. For video recordings, the privacy risks can be assessed using face verification systems and face reconstruction attacks\cite{Rosberg_2023_ICCV} to determine the degree of obfuscation. Synthetic data must be rigorously tested against membership inference attacks and attribute inference attacks \cite{goncalves2020generation,privacy-gain,MURTAZA2023100546,synthetic-privacy-all}. Outlier leakage is another critical concern, especially in mental health datasets, where outliers are more prevalent due to the diversity and small size of participant groups. Metrics such as privacy gain \cite{privacy-gain}, outlier similarity \cite{MURTAZA2023100546}, and reproduction rate \cite{MURTAZA2023100546} are effective for evaluating these risks. These tests should be applied across languages, demographics, and cultures, and must incorporate cross-modal privacy assessments to detect information leakage through combination of modalities. The empirical privacy measures should also be accompanied by theoretical guarantees similar to $\epsilon$-value in DP.
Utility evaluation should assess the usefulness of the data for mental health diagnosis tasks, focusing on information preservation \cite{SANCHEZ2014189,re-id}, intonation preservation \cite{VP2022}, conversational diversity \cite{VP2022,OHNN}, naturalness \cite{VP2022,multi-speaker-anon}, and emotional feature retention \cite{tomashenko2024voiceprivacy,vid_anon}. LLMs can be leveraged to automatically evaluate the utility of synthetically generated data on dimensions such as naturalness, empathy, helpfulness, professionalism, comprehensiveness, authenticity, and safety \cite{chen-etal-2023-soulchat,zhang2024cpsycounreportbasedmultiturndialogue} or psychological measures like the working alliance inventory \cite{qiu-2024}.

\paragraph{Privacy-aware model training.} During training, privacy-preserving methods like DP are critical. Mental health models either consist of models fine-tuned with mental health data or trained fusion layers to combine features extracted from various modalities using pre-trained models. To ensure privacy, these models should be fine-tuned with DP methods \cite{kerrigan-etal-2020-differentially, yu2022differentially}, and fusion layers should be trained using DP-SGD \cite{dp-sgd}. In multi-institutional settings, federated learning combined with local differential privacy can offer stronger protection \cite{NAGY2023110475,dp-fl1}.


\paragraph{Privacy-utility evaluation of privacy-aware training.}
Finally, the trained models must be evaluated for their privacy-utility trade-off. Privacy measurements include testing the models against membership inference attacks \cite{context-aware-dp-lm} and analyzing the theoretical guarantees provided by the $\epsilon$ value in differential privacy \cite{kerrigan-etal-2020-differentially,yu2022differentially,aud-dp,bu2022differentially}. To assess utility, the models should be evaluated on downstream mental health diagnosis tasks \cite{context-aware-dp-lm,plant-etal-2021-cape,kerrigan-etal-2020-differentially,yu2022differentially,bu2022differentially,ravuri,pranjal}. Additionally, testing on diverse datasets can help identify any biases or disparities amplified during the training process \cite{dp-sgd-bad-performance}.

\section*{Prospects}

\paragraph{Multi-Speaker Anonymization (MSA).} While Miao et al. \cite{multi-speaker-anon} provided a benchmark for MSA, they assume weak attack models where the attacker does not have knowledge about the used anonymization scheme or cannot untangle speakers in overlapping segments. In reality, attackers may possess knowledge of the anonymization strategies or have the ability to separate overlapping speech segments, posing severe privacy risks in therapy settings. Consequently, MSA methods must be evaluated under stronger threat models and MSA should be enhanced to ensure resilience against sophisticated attacks.


\paragraph{Anonymization in video.} Video anonymization techniques also remain limited. Even after face obfuscation, models often leak sensitive attributes such as gender and age \cite{human_attribute_privacy}. These attributes can be used to identify individuals especially in small mental health datasets with limited demographic representation. While recording the body language of patients could help in mental health diagnosis, it can also reveal the gender of the patient if only face anonymization is performed \cite{gender_leak_face_obfus}. Moreover, current methods are also prone to demographic unfairness \cite{demogarphic_face_privacy}. Thus it is essential to develop fair and improved video anonymization techniques that can prevent leakage of private information like age and gender.


\paragraph{Theoretical guarantees in data anonymization.} While we discuss various data anonymization processes for text, audio and video modalities, most of them do not provide any theoretical guarantees like DP provides in privacy-aware model training. In text modality, word-level or sentence-level perturbations through DP provide theoretical guarantees\cite{nap2}. However, they significantly reduce the utility of the text \cite{nap2}, necessitating better anonymization techniques with privacy guarantees for text. For audio, DP guarantees exist only in single-speaker anonymization settings \cite{shamsabadi:hal-03588932}, whereas therapy sessions involve multiple speakers. Extending DP-based anonymization to multi-speaker contexts remains an open challenge. In video, existing DP methods for images \cite{img-dp} and video object indistinguishability \cite{vid-dp} have not been adapted for privacy-preserving facial de-identification in therapy footage, calling for task-specific DP-based video anonymization techniques with validated utility and fairness. In synthetic data generation DP-based methods for theoretical guarantees have been explored for text \cite{yue-etal-2023-synthetic,nahid2024safesynthdp}, tabular data \cite{SUN2023104404,qian2024synthetic}, multimodal tabular and 3D image data \cite{ziegler2022multimodal} generation and with FL \cite{dp-fl-gan}. However, no DP-based methods have been developed for multimodal therapy session generation.

\paragraph{Multimodal data anonymization.} Multimodal data introduces unique privacy risks. Current anonymization methods treat each modality in isolation, but information can leak across modalities. For example, lip movements in video can reveal PII removed from text or audio. Addressing such vulnerabilities will require the development of multimodal adversarial re-identification models capable of identifying cross-modal risks and guiding the design of holistic anonymization techniques. This will require fundamentally new algorithms that jointly anonymize data across all modalities while preserving clinically relevant features.


\paragraph{Multimodal synthetic data generation.} Current multimodal synthetic data generation methods often lack psychiatric grounding, reducing their utility in mental health applications. Most rely solely on patient characteristics without integrating therapeutic frameworks like Cognitive Behavioral Therapy (CBT). Incorporating such models, akin to Patient-$\psi$\cite{patient-psi}, could enhance clinical relevance of the generated data. Additionally, existing approaches follow sequential pipelines: text generation followed by audio and video synthesis leading to inconsistencies and diminished authenticity. Leveraging advanced multimodal LLMs \cite{next-gpt} to generate cohesive, end-to-end therapy sessions provides a promising direction for improvement. However, current models struggle with coherence and memory in long multi-turn conversations \cite{multi-turn-struggle}, necessitating the development of long-context multimodal LLMs tailored to therapy data.


\paragraph{Privacy-utility evaluations for multimodal data and models.}
Despite progress in modality-specific privacy-utility evaluations, assessments for multimodal data remain underdeveloped. Cross-modal interactions may introduce novel vulnerabilities that require specialized evaluation frameworks. Additionally, comparative studies on the privacy-utility trade-offs between anonymized and synthetic data have yet to be conducted. Furthermore, privacy-preserving techniques often underperform across demographic groups, particularly those with distinctive facial traits \cite{demogarphic_face_privacy}, and differential privacy methods may exacerbate these fairness issues, amplifying biases against underrepresented populations \cite{dp-sgd-bad-performance}. Addressing these gaps requires evaluation must be conducted on datasets that reflect diverse demographics, languages, and cultures. This will ensure that privacy-preserving methods are inclusive, equitable, and effective across varied contexts.


\paragraph{Local Differential Privacy for Federated Learning (LDP-FL).} While LDP-FL has been explored in recent times \cite{NAGY2023110475,dp-fl1}, there has not been enough privacy-utility evaluation in mental health tasks. Basu et al. \cite{dp-fl-benchmark} demonstrated that LDP-FL experiences greater utility degradation when applied to realistic data resembling medical datasets, small datasets, or large models. Therefore, a LDP-FL setup with an improved privacy-utility trade-off is necessary. Additionally, no existing work compares the privacy performance of DP methods with LDP-FL methods, a comparison essential for determining the most suitable approach.

\section*{Conclusion}

This paper outlines key challenges in building AI systems for mental health diagnosis, emphasizing the privacy risks inherent in handling sensitive data. We critically reviewed core strategies to mitigate these risks, including text, voice, and face anonymization; synthetic data generation that replicates real-world interactions without revealing personal information; and privacy-preserving model training using differential privacy. Additionally, we detailed evaluation frameworks to assess the privacy and utility trade-offs of these methods to ensure clinical relevance is maintained while safeguarding confidentiality.
Building on this, we proposed a comprehensive pipeline for developing privacy-aware mental health AI models, encompassing data collection, anonymization, synthetic data generation, privacy-utility evaluations, and privacy-aware training. This workflow aims to balance privacy protection with the utility required for effective mental health diagnosis and therapy assistance. 
Looking ahead, we highlight critical research directions: developing anonymization techniques that address cross-modal privacy risks, generating clinically grounded multimodal synthetic data, and establishing inclusive evaluation protocols across demographics and cultures. These advances are essential for deploying trustworthy, privacy-preserving AI systems that can expand access to mental health care while upholding the highest standards of data privacy and security.

\section*{Author Contributions} I.G., A.M. and T.C. contributed to conceptualizing the manuscript. A.M. led the effort of writing the initial draft of the manuscript. I.G., A.M. and T.C. finalized the manuscript. T.C. and I.G. supervised the project.

\section*{Funding Information}
This research work of I.G. and A.M. has been supported by the German Federal Ministry of Education and Research and the Hessian Ministry of Higher Education, Research, Science and the Arts within their joint support of the National Research Center for Applied Cybersecurity ATHENE, by the LOEWE Chair of Research Excellence “Ubiquitous Knowledge Processing” (Grant Number: LOEWE/4a//519/05/00.002(0002)/81)and by the DYNAMIC center, both of the former funded by the LOEWE program of the Hessian Ministry of Science and Arts (Grant Number: LOEWE/1/16/519/03/09.001(0009)/98). T.C. acknowledges travel support from the European Union’s Horizon 2020 research and innovation programme under Grant Agreement No 951847 and Rajiv Khemani Young Faculty Chair Professorship in Artificial Intelligence and  the support of the Alexander von Humboldt Foundation through a Humboldt Research Fellowship for Experienced Researchers.

\section*{Competing Interests}
The authors declare no competing interests.

\section*{Additional Information}

\noindent{\bf Materials \& Correspondence} should be emailed to Tanmoy Chakraborty (\url{tanchak@iitd.ac.in}) and Iryna Gurevych (\url{iryna.gurevych@tu-darmstadt.de}).

\bibliography{sample}

\end{document}